\documentclass{article}
\usepackage[utf8]{inputenc}
\usepackage{graphicx}
\usepackage{amsmath}
\usepackage{hyperref}
\usepackage{geometry}
\usepackage{enumitem}
\usepackage{listings}
\usepackage{tcolorbox}
\tcbuselibrary{listingsutf8}
\usepackage{verbatim}
\usepackage{tikz}
\usetikzlibrary{shapes.geometric, arrows}
\usepackage{multirow}
\usepackage{booktabs}
\usepackage{listings}
\usepackage{xcolor}
\definecolor{lemonbg}{RGB}{255, 252, 210}
\usepackage{listings}
\usepackage{courier}

\title{Task Memory Engine (TME): Enhancing State Awareness for Multi-Step LLM Agent Tasks}
\author{Ye Ye \\}
\date{April 2025}

\begin{document}

\maketitle

\begin{abstract}
Large Language Models (LLMs) are increasingly used as autonomous agents for multi-step tasks. However, most existing frameworks fail to maintain a structured understanding of the task state, often relying on linear prompt concatenation or shallow memory buffers. This leads to brittle performance, frequent hallucinations, and poor long-range coherence. In this work, we propose the \textbf{Task Memory Engine (TME)}, a lightweight and structured memory module that tracks task execution using a hierarchical Task Memory Tree (TMT). Each node in the tree corresponds to a task step, storing relevant input, output, status, and sub-task relationships. We introduce a prompt synthesis method that dynamically generates LLM prompts based on the active node path, significantly improving execution consistency and contextual grounding. Through case studies and comparative experiments on multi-step agent tasks, we demonstrate that TME leads to better task completion accuracy and more interpretable behavior with minimal implementation overhead.
A reference implementation of the core TME components is available at \href{https://github.com/biubiutomato/TME-Agent}{https://github.com/biubiutomato/TME-Agent}, including basic examples and structured memory integration.
While the current implementation uses a tree-based structure, TME is designed to be graph-aware, supporting reusable substeps, converging task paths, and shared dependencies. This lays the groundwork for future DAG-based memory architectures.
\end{abstract}

\vspace{1em}  
\vspace{1.5em}  

\section{Introduction}

LLM-based agents have been widely adopted across domains such as customer support, automation, and complex task planning. However, these agents often struggle with maintaining coherent task context over time, especially in multi-turn or multi-step workflows where precise planning and execution are crucial. Traditional memory mechanisms, such as linear history or basic vector retrieval, fail to capture the complexity of task execution. Linear history lacks structural awareness, leading to irrelevant or redundant information in prompts, while vector retrieval misses critical temporal dependencies and context-specific nuances. As a result, there is no standardized method for managing an LLM agent's internal task state, resulting in inefficiencies and suboptimal performance, particularly in long-horizon planning tasks.

We argue that \textbf{structured state modeling} is essential for robust task execution and long-horizon planning. To address this need, we propose the \textbf{Task Memory Engine (TME)}, a novel memory architecture composed of three core components: the \textbf{Task Memory Tree (TMT)}, the \textbf{Task Relationship Inference Module (TRIM)}, and the \textbf{Prompt Synthesizer}.

TMT provides a hierarchical representation of task progression, enabling state tracking and recovery. TRIM performs fine-grained reasoning over task relationships—such as dependency, rollback, and merging—using rule-based logic, cosine similarity, and a transformer-based classification module. The Prompt Synthesizer generates adaptive context prompts based on the active path in TMT.

with future support planned for GNN-based and LLM-augmented inference.

Together, these components enable robust, interpretable, and token-efficient execution of complex, multi-step tasks.

Our main contributions are as follows:
\begin{enumerate}[label=\bfseries\arabic*.]
    \item We introduce \textbf{Task Memory Engine (TME)}, a novel memory framework composed of the \textbf{Task Memory Tree (TMT)} and the \textbf{Task Planner}, which together support structured, hierarchical, and revisitable task execution and planning for LLM agents.
    
    \item We present the \textbf{Task Memory Tree (TMT)}, a lightweight internal representation designed for efficient state tracking, backtracking, and loop-aware reasoning in task planning.
    
    \item We introduce the \textbf{Task Relationship Inference Module (TRIM)}, a symbolic reasoning module that supports rollback, dependency management, subtree merging, and control flow revision in real time.
    
    \item We present the \textbf{Prompt Synthesizer}, a lightweight component that generates compact, context-aware prompts from the task execution path in TMT, enabling memory-efficient and coherent interactions.
    
    \item We validate TME’s effectiveness through \textbf{realistic case studies and token efficiency comparisons}, demonstrating significant improvements over traditional linear prompt chaining, especially in task planning scenarios.
    
    \item We release an \textbf{open-source implementation of TME and the Task Planner}, making them accessible for a wide range of agent-based applications, particularly those involving complex task planning and execution.
\end{enumerate}

\begin{figure}[htbp]
  \centering
  \includegraphics[width=0.6\linewidth]{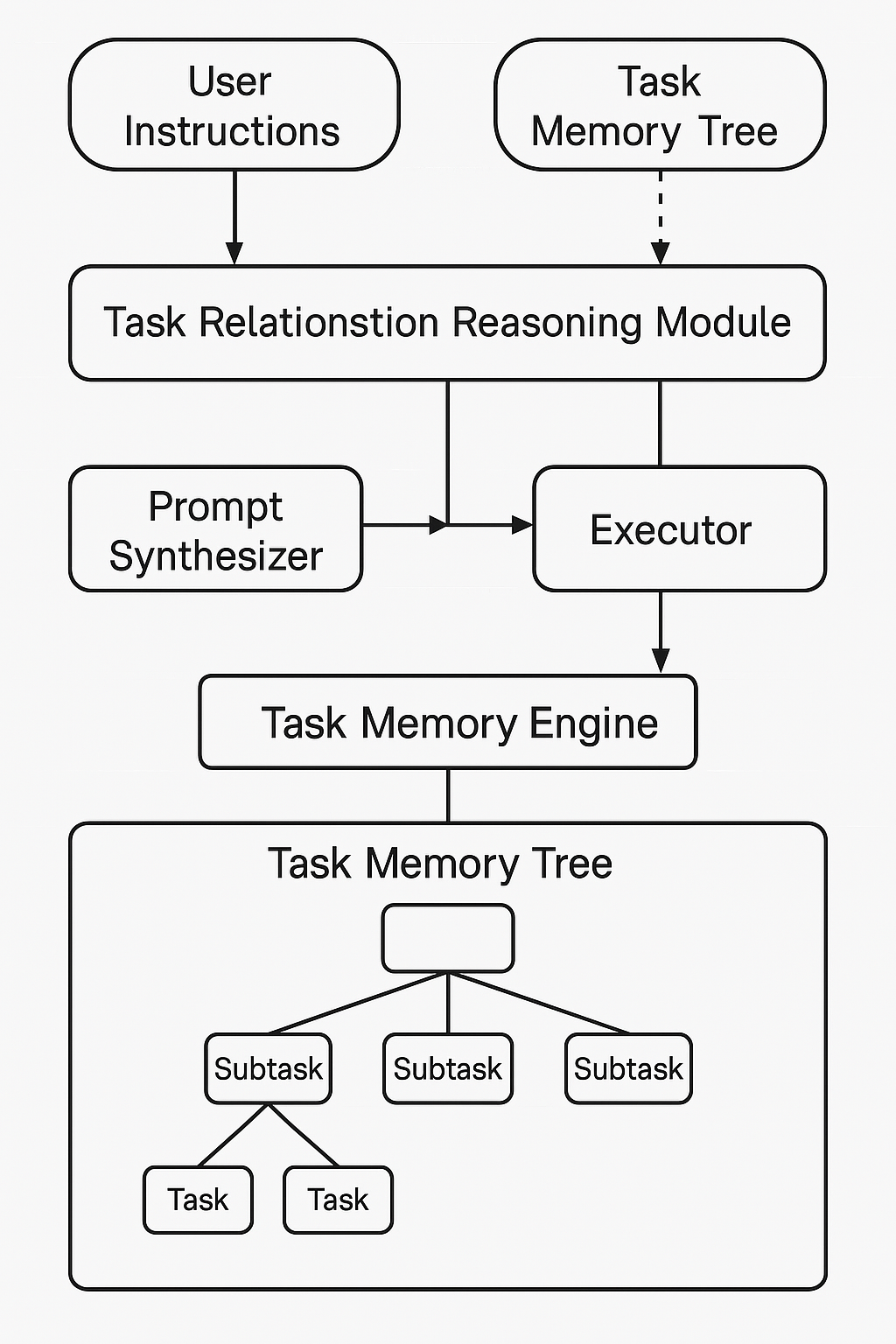}
  \caption{Overview of the Task Memory Engine (TME) Architecture. The system consists of a Task Memory Tree (TMT), a Task Relationship Reasoning Module, a Prompt Synthesizer for executing subtasks, and an Execution Feedback Loop that supports rollback and dynamic updates.}
  \label{fig:tme-architecture}
\end{figure}

\section{Related Work}

Recent research has explored various forms of state modeling for long-horizon or multi-step task execution with LLMs. For example, \textbf{Stateflow} models LLM workflows as finite-state machines to support state-driven execution policies~\cite{wu2024stateflow}. \textbf{LLM-State} focuses on external world state representation for planning, particularly in open-world settings~\cite{chen2023llmstate}. \textbf{STEP} introduces a stack-based policy model for hierarchical web actions~\cite{sodhi2023step}. \textbf{Dynamic Planning with an LLM} leverages symbolic planners with LLM agents for goal-oriented decision making~\cite{dagan2023dynamic}.

While these works contribute to planning and state abstraction, few provide memory structures designed specifically for managing prompt-level execution in interactive agents~\cite{wu2024stateflow, chen2023llmstate, dagan2023dynamic}. In contrast, our approach focuses on \textbf{internal, structured, and revisitable task memory}, suitable for LLMs that must support non-linear workflows, backtracking, and task recovery.

We review several prominent approaches to LLM memory and agent architectures:

\begin{itemize}
    \item \textbf{Prompt-based memory} (e.g., LangChain's \texttt{BufferMemory}): simply appends previous interactions.
    \item \textbf{Tool execution logs} (e.g., \texttt{AutoGPT}~\cite{auto-gpt}, \texttt{BabyAGI}): track tool calls but not logical task progression.
    \item \textbf{Vector stores} (e.g., Pinecone, Weaviate): allow retrieval but do not model current state.
    \item \textbf{ReAct-style prompting}~\cite{yao2022react}: combines reasoning and action but has no memory persistence.
\end{itemize}

None of these approaches provide a formal representation of the task state tree, and few can support task recovery or task-level inspection.

\section{Problem Definition}

We define the \textbf{task state} as the structured representation of an agent's progress through a multi-step objective. This includes:

\begin{itemize}
    \item Current step and status (waiting, active, done)
    \item Parent-child relationships between steps (for sub-task modeling)
    \item Inputs, outputs, and results for each step
    \item Metadata such as timestamps, errors, or retry attempts
\end{itemize}

A system without such modeling risks losing track of where it is in the task, making it prone to redundant actions or failure.

\section{Method: Task Memory Engine}

The Task Memory Engine (TME) is a modular and efficient memory framework that enhances task reasoning for LLM-based agents through three tightly integrated components: the \textbf{Task Memory Tree (TMT)}, the \textbf{Task Relationship Inference Module (TRIM)}, and the \textbf{Prompt Synthesizer}. TMT tracks task structure and state. TRIM interprets and manages task relationships such as dependencies, rollbacks, and merges. The Prompt Synthesizer leverages the active path in TMT to generate minimal yet coherent prompts for each step.

We propose the \textbf{Task Memory Engine (TME)}, a lightweight and modular memory layer designed to improve multi-step task reasoning in LLM-based agents. TME enhances state tracking, enables contextual prompt construction, and supports backtracking and branching workflows.

\subsection{Task Memory Tree (TMT)}

The \textbf{Task Memory Tree (TMT)} is a hierarchical data structure designed to support consistent and efficient task tracking for large language model (LLM) agents. Each node in the tree represents a single step in a task and maintains structured metadata that enables non-linear reasoning, memory-efficient prompt synthesis, and dynamic workflow management.

\paragraph{Node Structure.} 
Each node in the TMT corresponds to a discrete task step and contains the following information:
\begin{itemize}
    \item \textbf{Action:} The operation being performed, such as ``Search for flights'' or ``Compare prices''.
    \item \textbf{Input/Output:} Associated data including user queries, system responses, or intermediate results.
    \item \textbf{Status:} The current execution state of the task step (e.g., active, completed, failed).
    \item \textbf{Parent/Children:} Explicit hierarchical relationships enabling traversal and subtask nesting.
    \item \textbf{Dependencies:} Optional cross-links that describe logical relations among tasks across branches.
\end{itemize}

\paragraph{Dynamic Execution and Prompt Synthesis.}
As an agent progresses through a task, nodes are dynamically created and updated in real time. A dedicated \textit{Prompt Synthesizer} traverses the active node path — from the root to the current leaf — and generates a concise, contextually-relevant prompt. This enables:
\begin{itemize}
    \item Efficient memory usage by limiting the prompt to relevant history,
    \item Reduced token overhead during inference,
    \item Improved coherence of agent behavior across complex and long-running tasks.
\end{itemize}

\paragraph{Support for Non-linear and Reusable Workflows.}
TMT is optimized for advanced task flows, supporting features such as:
\begin{itemize}
    \item \textbf{Revisitable branches:} Nodes may store alternate execution paths, functioning as logical checkpoints.
    \item \textbf{Loop handling:} Cyclic behaviors are modeled via loop references and snapshot-based state tracking.
    \item \textbf{Subtree pruning and merging:} Heuristics are used to reduce redundancy, such as merging semantically similar branches (e.g., multiple “Add tomato” actions) or pruning irrelevant paths.
\end{itemize}

These capabilities allow the TMT to manage complex workflows like recipe adaptation, travel itinerary revisions, or multi-tab planning while maintaining clarity and consistency. We provide a serializable JSON schema for TMT and demonstrate its integration into an LLM control loop, making it a practical solution for memory-augmented agents.

While TMT is implemented as a tree structure, certain workflows (e.g., multi-tab reuse, loop shortcuts, or subtree merges) may induce implicit DAG-like patterns. These characteristics suggest that TMT is already approaching a relaxed DAG form, which lays the foundation for future graph-based memory extensions.

\subsection{Task Relationship Inference Module (TRIM)}

To enhance structural reasoning and flexible control flow, the \textbf{Task Relationship Inference Module (TRIM)} serves as the central logic engine within TME. Built atop the Task Memory Tree (TMT), TRIM maintains a structured representation of ongoing task execution, enabling efficient navigation, rollback, and prompt construction for long or complex workflows.

\subsubsection{Structured Task State Tracking}
Each task step is explicitly recorded as a node in the Task Memory Tree, with annotated parent-child relationships, execution status, and content. This structure allows the agent to infer which steps are active, completed, or invalidated without needing to resend the entire history.

\subsubsection{Relationship Classification}
TRIM infers structural relationships between user instructions and existing nodes to update the TMT accordingly. Five core task relationships are supported:
\begin{itemize}
  \item \textbf{Dependency} (\texttt{depends\_on}): Task B requires Task A to complete first.
  \item \textbf{Parallel} (\texttt{parallel\_with}): Task A and B can proceed independently.
  \item \textbf{Replacement} (\texttt{replaces}): Task B overrides a previous plan or step.
  \item \textbf{Merge} (\texttt{merge}): Task B and Task A merge to one task.
  \item \textbf{Rollback} (\texttt{rollback}): Task A failed or was canceled; revert to an earlier point.
  \item \textbf{Subtask} (\texttt{child\_of}): Task B is a decomposed subgoal of Task A.
\end{itemize}
These relationships are currently inferred using a hybrid of prompt templates, keyword matching, and logic rules, with future support planned for a lightweight learned classifier. To support this, TRIM computes \textbf{cosine similarity over sentence embeddings} between the new instruction and existing task nodes. High semantic similarity combined with matching execution state can trigger rollback or replacement; moderate similarity suggests dependencies; low similarity implies a parallel or new subtask. These inferences are further refined through lightweight rule-based constraints.

\subsubsection{Structure-Aware Relationship Inference}
We currently implement relationship classification using:
\begin{itemize}
  \item Embedding-based cosine similarity to detect task similarity.
  \item Tree-based structural update rules to determine parent-child or rollback links.
\end{itemize}

These approaches allow TRIM to identify surface-level repetition and local dependencies. However, they may fail in semantically equivalent but lexically divergent instructions, or in multi-turn workflows where task intents evolve gradually.

\subsubsection{Enhanced Inference via Lightweight Relationship Classifier}

To improve inference beyond similarity heuristics, we introduce a lightweight classification module for task relationship prediction. Given a pair $(S, N_i)$ where $S$ is the current step and $N_i$ is a historical node, the classifier outputs a relationship type:

\[ R_{S,N_i} \in \{ \texttt{duplicate}, \texttt{superset}, \texttt{depends\_on}, \texttt{conflict}, \texttt{unrelated} \} \]

The classifier is implemented using a compact transformer encoder (e.g., MiniLM or DistilBERT) followed by a multi-class softmax layer. Input features include the raw natural language descriptions of $S$ and $N_i$, optionally augmented by structured node metadata (e.g., status, subtree hash).

This model can be pre-trained on synthetic task graphs or fine-tuned on labeled multi-step interaction traces. Once deployed, it enables:
\begin{itemize}
  \item Precise task node merging (e.g., semantic duplication elimination)
  \item Dependency-aware updates (e.g., delay execution of dependent nodes)
  \item Conflict detection and rollback routing
\end{itemize}

By integrating this module, TRIM moves beyond static heuristics and acquires adaptive relational reasoning capabilities necessary for robust long-horizon task planning.

\subsubsection{Future Directions: Beyond Classification}

While the lightweight classifier enhances local relational reasoning, it remains limited in capturing global task context or compound interactions. To further strengthen TRIM's inference capabilities, we envision several extensions:

\begin{itemize}
  \item \textbf{Span-based Coverage Models}: Instead of discrete labels, predict whether a current step semantically covers a subtree (e.g., nodes $N_3$ to $N_5$) and the confidence of such coverage.
  \item \textbf{Graph Neural Networks}: Model the entire Task Memory Tree as a dynamic graph, enabling relational reasoning across multiple paths, dependencies, and hierarchy levels.
  \item \textbf{LLM-Augmented Reasoning}: Use prompt-based large language models to perform high-level judgment when task overlaps are fuzzy, ambiguous, or involve contextual paraphrasing.
\end{itemize}

These strategies would enable TRIM to go beyond static pairwise prediction and move toward a globally consistent, structure-aware memory inference engine.

\subsubsection{Non-linear Navigation and Rollback}
TRIM supports flexible workflows through non-linear navigation. Users may revisit and edit previous steps (e.g., correcting mistakes or changing preferences). Only affected subtrees are updated, avoiding full sequence recomputation. This makes TRIM particularly suitable for agents operating in dynamic environments.

\subsubsection{Task Merging and Pruning}
When redundant or semantically similar task branches occur, TRIM merges them into unified nodes to minimize duplication:
\[
\text{Merged Task} = f(\text{Task}_1, \text{Task}_2)
\]
Branches that are invalidated or unreachable are pruned to minimize execution and prompt overhead:
\[
\text{Pruned Tree} = \arg \min \left( \sum_{i=1}^{n} \text{cost}_i \right)
\]

\subsubsection{Token-Efficient Prompt Synthesis}
Traditional agents pass the full dialogue history to the model at each step, leading to token inefficiency. TRIM cooperates with the Prompt Synthesizer to extract only the relevant node path (from root to current leaf):
\[
\text{Optimal Path} = \arg \min_{p \in P} \left( \sum_{i=1}^{n} \text{cost}_i \right)
\]
This ensures minimal token use while maintaining logical coherence and contextual fidelity.

\subsubsection{Key Functions Summary}
\begin{itemize}
    \item Determine parent-child and hierarchical task relations
    \item Detect and merge redundant or equivalent task steps
    \item Infer sequential dependencies and resolve conflicts
    \item Support rollback, re-entry, and alternative path creation
\end{itemize}

\textbf{Section Summary:} TRIM enables symbolic task introspection, dynamic execution correction, and efficient planning for real-world agent workflows—particularly when long-term memory, partial updates, or multi-path execution are involved.

\vspace{0.5em}
\begin{figure}[ht]
\centering
\begin{minipage}{0.85\linewidth}
\begin{verbatim}
Fill Form
|------ Name
|   |------ Initial: John Doe
|   |------ Correction: John Smith   (rollback)
|------ Email
|------ Submit
\end{verbatim}
\end{minipage}
\caption{A rollback-aware task structure. The initial name “John Doe” is later corrected to “John Smith” and merged into the task node.}
\label{fig:rollback_tree}
\end{figure}

\subsection{Integration of TMT and TRIM}
Together, TMT and TRIM form a cohesive framework that enables the agent to execute complex tasks with better memory efficiency, coherence, and flexibility. While TMT provides the structural representation of the task space, TRIM manages the relationships, dependencies, and dynamic changes within that space.


\begin{itemize}
    \item \textbf{Dynamic Backtracking and Branching}: TMT allows for flexible task navigation, while TRIM ensures that the task is backtracked, adjusted, and re-executed efficiently as needed. This combination supports complex branching and re-evaluations that are common in real-world task planning (e.g., revisiting decisions or adapting to new inputs).
    \item \textbf{Efficient Task State Management}: TRIM enhances the task management process by ensuring that each task step is updated with minimal overhead, allowing for scalable task execution across multiple agents or steps without redundantly reprocessing earlier parts of the task.
    \item \textbf{Scalability and Flexibility}: The integration allows the system to scale effectively with increasing task complexity, accommodating parallel subtask execution, nested workflows, and long-term task objectives. By maintaining a structured representation and providing an inference mechanism, TME supports both short-term task execution and long-term planning with minimal overhead.
\end{itemize}

\subsection{The Role of Prompt Synthesizer}
A crucial part of the TME architecture is the \textbf{Prompt Synthesizer}, which dynamically constructs relevant prompts based on the active node path in the TMT. By selecting only the necessary parts of the task history, the synthesizer ensures that the prompts are token-efficient, avoiding unnecessary repetitions and redundancies. This approach helps reduce the token costs associated with maintaining long, complex task histories and guarantees that the agent operates within memory and token usage constraints.

\subsection{Task Planner Module (Optional Extension)}

To support long-horizon workflows and proactive goal decomposition, the Task Memory Engine (TME) can optionally integrate a Task Planner module. This component is designed to operate at a higher abstraction level than TRIM, focusing on generating the initial task structure and adapting it in response to updated goals or feedback.

\textbf{Planner Responsibilities.} The planner serves as a front-end coordinator for high-level objectives, producing a hierarchical task outline that populates the Task Memory Tree (TMT). Key responsibilities include:
\begin{itemize}
    \item \textbf{Goal Decomposition:} Translating user intent or high-level prompts into structured subgoals using LLM prompting or symbolic methods.
    \item \textbf{Action Proposals:} Suggesting candidate action steps or tool invocations that can fulfill subgoals.
    \item \textbf{Sequencing:} Estimating logical or temporal dependencies among subgoals, possibly through pattern-based rules or vector similarity.
\end{itemize}

\textbf{Integration with TRIM.} Once the initial task structure is proposed, TRIM may be invoked to refine the relationships between tasks. For instance, the planner may generate a linear list of subtasks, which TRIM re-organizes into a dependency-aware tree or graph using rollback, merge, and parallelization heuristics.

\textbf{Planning Techniques.} The planner may employ several strategies depending on system capability and task domain:
\begin{itemize}
    \item \textbf{LLM-based Prompt Templates:} Recursive prompting patterns to iteratively expand each task node.
    \item \textbf{Classical Planning Algorithms:} Integration with STRIPS, PDDL, or cost-based planning engines.
    \item \textbf{Few-shot Sketching:} Using demonstrations of task decompositions to bootstrap structural predictions.
\end{itemize}

\textbf{Future Extensions.} Future versions of the planner may support:
\begin{itemize}
    \item \textbf{Planner-LLM Co-design:} Co-training or co-prompting with TRIM for tighter semantic alignment.
    \item \textbf{Goal Re-planning:} When user intent changes, re-evaluate and restructure downstream subgoals.
    \item \textbf{Traceable Goal Trees:} Annotated visual trees that support debugging and editing of task flow.
\end{itemize}

This optional module complements TME by introducing a proactive and structured approach to goal translation, enabling agents not only to reactively manage task memory but also to plan ahead with semantic awareness.

\subsection{Section Summary}
The integration of TMT and TRIM in the Task Memory Engine (TME) provides a powerful and flexible framework for multi-step task execution and planning. By combining hierarchical state tracking, dynamic path re-routing, and efficient prompt construction, TME enables agents to handle complex workflows with improved efficiency, coherence, and memory usage. The system is well-suited for tasks that require dynamic task execution, non-linear paths, corrections, and re-evaluations, making it highly adaptable to a wide range of applications in agent-based task planning and execution.

\section{Case Study: Form-Filling Assistant}

We simulate a multi-turn form-filling task using the Task Execution Tree (TET) approach. The task involves collecting, correcting, and confirming user information over six interaction rounds.

\textbf{Input Sequence:}
\begin{lstlisting}
// ~~~~ Round 1 ~~~~
Question: Help me fill out a form, I will provide some of my information to you.
Answer: Of course! Please provide me with the information you would like me to include in the form.

// ~~~~ Round 2 ~~~~
Question: My name is John Doe.
Answer: Great, thank you! What other information would you like me to include in the form.

// ~~~~ Round 3 ~~~~
Question: My email is john@example.com.
Answer: Great, thank you for sharing your email address. Can you please provide me with the rest of the information needed to fill out the form.

// ~~~~ Round 4 ~~~~
Question: My address is Market Street, San Francisco.
Answer: Great, thank you for providing your address. Can you please provide me with your full name, date of birth, email address, and phone number so I can complete the form accurately.

// ~~~~ Round 5 ~~~~
Question: Sorry, to correct, my name is John Smith.
Answer: Got it, thank you for the correction. Please provide me with the rest of the information you would like me to include in the form.

// ~~~~ Round 6 ~~~~
Question: Help to repeat my information, Then submit.
Answer: Sure! Here is the information you provided:
Name: John Smith
Email: john@example.com
Address: Market Street, San Francisco
I will now submit the form with this information. Thank you for your assistance!
\end{lstlisting}

The form-filling task is structured into six logical nodes, including information collection (name, email, address), a correction step for name, and a final submission. Rather than duplicating corrected input, the Task Execution Tree merges revisions into their respective nodes, enabling efficient updates while preserving history.

The task tree enables correction and re-submission without replaying the full interaction history, maintaining clean and compact context.

\begin{figure}[ht]
\centering
\includegraphics[width=0.9\linewidth]{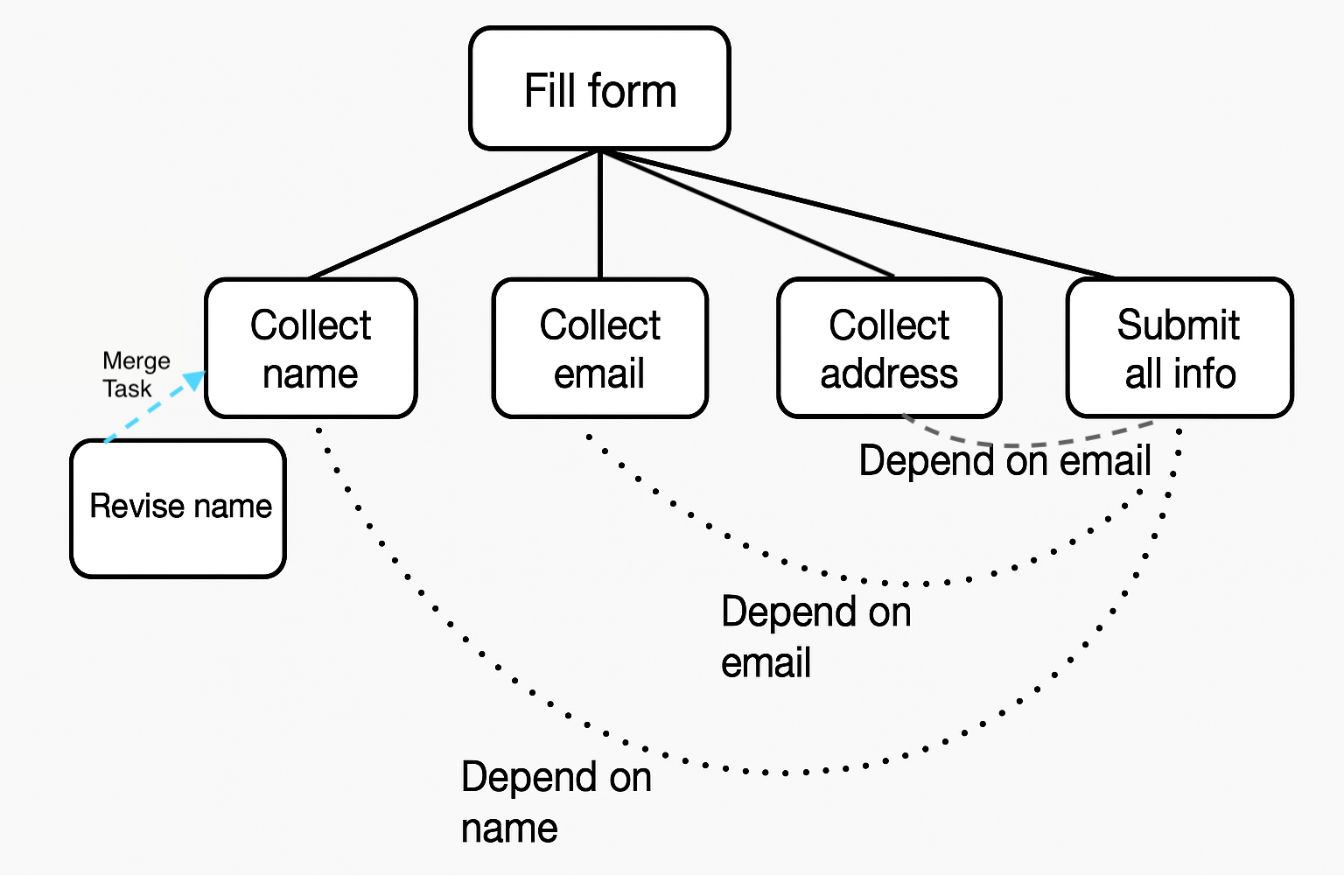}
\caption{
Task Execution Tree (TET) representation for a multi-turn form-filling assistant. 
Each leaf node corresponds to a specific field collection (name, email, address), while dotted arrows indicate inter-step dependencies. The “Revise name” task is merged into the original “Collect name” node. 
The final “Submit all info” node depends on the successful completion of upstream fields.
}
\label{fig:fillform_tet}
\end{figure}

Figure~\ref{fig:fillform_tet} illustrates the Task Execution Tree (TET) for a structured form-filling assistant. The root node \texttt{Fill form} governs the overall task. Each sub-node represents a field-specific step: collecting the user's name, email, or address. Dependencies between tasks (e.g., \texttt{Submit all info} depends on \texttt{email}, \texttt{name}, etc.) are shown with dotted arrows, modeling how downstream tasks require upstream inputs.

A correction task such as \texttt{Revise name} is merged into the existing \texttt{Collect name} node via a \textit{Merge Task} operation. Rather than creating a new branch or replacing the original input, the TET consolidates both the original and revised prompts into a unified node, preserving the complete history related to the user's name. This enables precise revision tracking while avoiding the need to replay unrelated context. The tree structure thus supports modular updates, dependency reasoning, and efficient context reconstruction.

\subsection*{Token Efficiency Analysis}

The majority of token savings in TET stem from its prompt construction strategy. Unlike the Baseline method, which appends the entire conversation history at each round, TET references only the relevant task node and memory state. This design avoids redundant prompt repetition and enables highly efficient multi-turn interaction. See Appendix Table~\ref{tab:prompt_construction_comparison} for detailed per-round prompt content.

To evaluate efficiency, we compare TET with a baseline that naively appends full history at each round. Tables~\ref{tab:token_usage_prompt_completion_total} and~\ref{tab:token_usage_comparison_table} summarize the prompt, completion, and total token usage.

\begin{table}[h]
\centering
\begin{tabular}{|c|cc|cc|cc|}
\hline
\textbf{Round} &
\multicolumn{2}{c|}{\textbf{Prompt Tokens}} &
\multicolumn{2}{c|}{\textbf{Completion Tokens}} &
\multicolumn{2}{c|}{\textbf{Total Tokens}} \\
\cline{2-7}
& Baseline & TET & Baseline & TET & Baseline & TET \\
\hline
1 & 24 & 24 & 25 & 25 & 49 & 49 \\
2 & 62 & 63 & 18 & 19 & 80 & 82 \\
3 & 94 & 64 & 22 & 24 & 116 & 88 \\
4 & 132 & 66 & 32 & 38 & 164 & 104 \\
5 & 182 & 100 & 33 & 23 & 215 & 123 \\
6 & 231 & 231 & 44 & 48 & 275 & 279 \\
\hline
\end{tabular}
\caption{Per-round token usage across Prompt, Completion, and Total tokens for Baseline and TET. TET reduces redundancy through structured memory reuse.}
\label{tab:token_usage_prompt_completion_total}
\end{table}

\begin{figure}[h]
\centering
\includegraphics[width=0.85\linewidth]{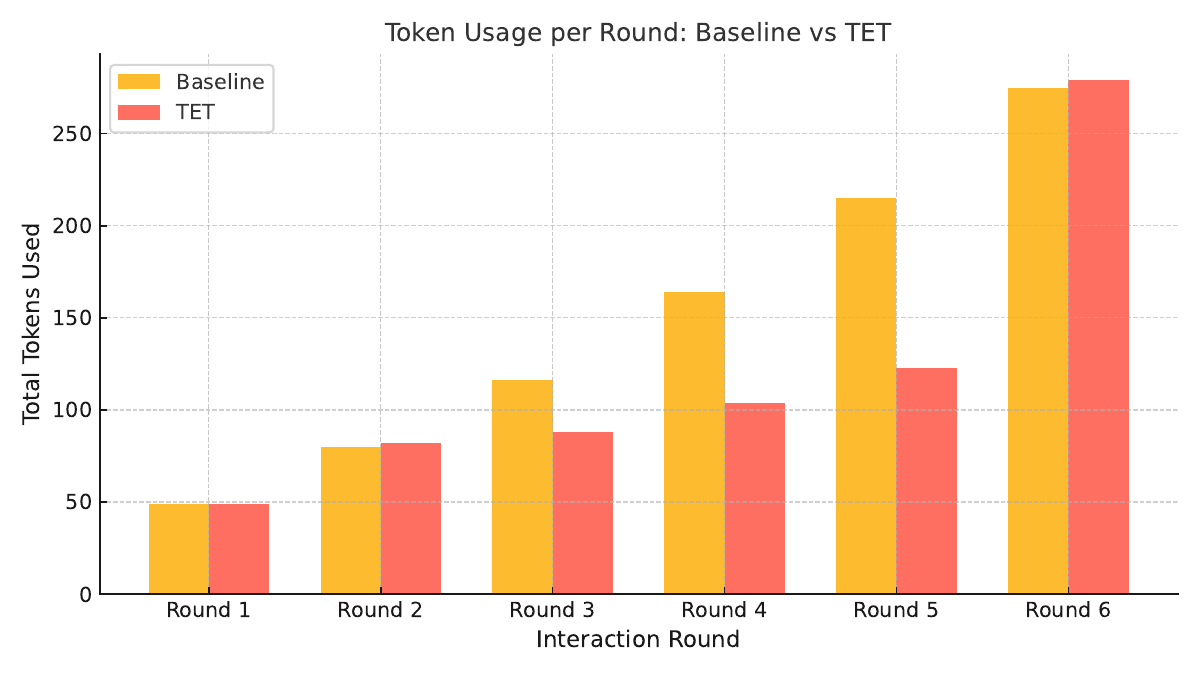}
\caption{Token usage comparison across 6 interaction rounds. The Baseline method grows linearly due to full prompt replay, while TET minimizes prompt length using task memory.}
\label{fig:formfilling_token_usage_comparison}
\end{figure}

\begin{table}[h]
\centering
\begin{tabular}{@{}lccc@{}}
\toprule
\textbf{Round} & \textbf{Baseline Tokens} & \textbf{TET Tokens} & \textbf{Tokens Saved} \\
\midrule
Round 1        & 49                      & 49                 & 0                     \\
Round 2        & 80                      & 82                 & -2                    \\
Round 3        & 116                     & 88                 & 28                    \\
Round 4        & 164                     & 104                & 60                    \\
Round 5        & 215                     & 123                & 92                    \\
Round 6        & 275                     & 279                & -4                    \\
\midrule
\textbf{Total} & \textbf{899}            & \textbf{725}       & \textbf{174}          \\
\bottomrule
\end{tabular}
\caption{Total token usage and savings per round. TET saves 174 tokens (19.4\%) across 6 rounds, with major reductions in mid-task updates.}
\label{tab:token_usage_comparison_table}
\end{table}

\begin{figure}[h]
\centering
\includegraphics[width=0.65\linewidth]{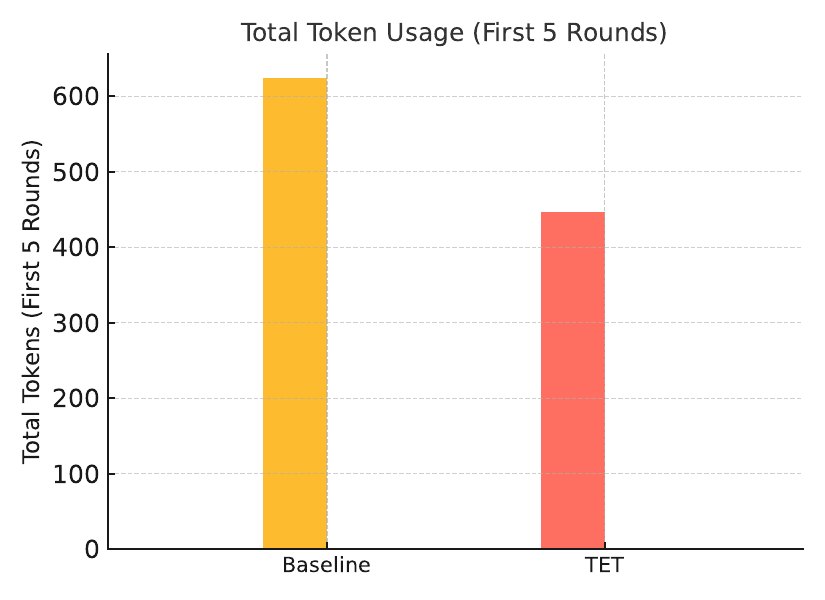}
\caption{
\textbf{Token usage comparison for the first 5 rounds.}
TET saves 178 tokens (26.4\%) by avoiding prompt replay during mid-task state edits. This highlights TET's advantage in tasks requiring corrections or dynamic updates.
}
\label{fig:token_usage_5rounds}
\end{figure}

\textbf{Insight:} TET reduces token usage by 174 tokens (19.4\%) over 6 rounds, and by 178 tokens (26.4\%) over the first 5 rounds. These savings arise from avoiding full-history repetition, especially during corrections and re-queries.

\textbf{Advantages of TET:}
\begin{itemize}
    \item Avoids redundant prompt concatenation
    \item Maintains task consistency across updates
    \item Enables token-efficient multi-turn interaction
\end{itemize}

\section{Reference Implementation}

We release a reference implementation of \textbf{TME} with task tree tracking, prompt synthesizer, and planner components. The community is invited to extend this abstraction layer to copilots, workflow agents, and personalized planners.

\textbf{Benefits:}
\begin{itemize}
    \item Token-efficient revision and re-planning
    \item Clear branch switching with preserved history
    \item No redundant re-evaluation of obsolete tasks
\end{itemize}

\vspace{0.5em}

\section{Discussion: Tree vs. DAG Memory Structures}

While the current implementation uses a tree-based memory structure (TMT), more general task execution workflows may naturally exhibit graph-like patterns. For example, the same subtask (e.g., "Verify user identity") might be referenced by multiple parent tasks. A Directed Acyclic Graph (DAG) representation enables:

\begin{itemize}
    \item \textbf{Shared subtask reuse} without duplication
    \item \textbf{Multi-path dependency resolution} across branches
    \item \textbf{Fine-grained credit assignment} in plan evaluation
\end{itemize}

However, DAG-based memory introduces complexity in state tracking and rollback reasoning. In contrast, trees offer a simpler model with deterministic navigation and local updates.

Future extensions of the TME framework may incorporate a DAG-backed memory structure to support reusable modules and convergent workflows. A comparative analysis of memory efficiency and computational cost between TMT and DAG variants is left as future work.

\section{Conclusion \& Future Work}

We present the \textbf{Task Memory Engine (TME)}, a structured memory framework for LLM-based agents that enables multi-step task planning, rollback, and preference-aware adaptation through a tree-based execution memory. By maintaining a hierarchical representation of task states, TME improves consistency, interpretability, and recovery in long-horizon agent workflows.

Our case studies demonstrate that TME outperforms baseline memory schemes in handling user preference updates, constraint violations, and multi-turn task coherence, while also reducing token overhead via structured prompt synthesis.

\textbf{Future directions include:}
\begin{itemize}
    \item Extending tree-based memory into DAG structures to model cross-step dependencies
    \item Integrating visual debugging tools to inspect and replay task flows
    \item Embedding TME into ReAct/AutoGPT-style planning loops for real-time adaptation
    \item Designing a planner module for long-term goal decomposition
    \item Supporting loop modeling, subtree merging, and version-aware memory navigation
    \item Systematic comparison with alternative memory paradigms beyond tree/DAG
\end{itemize}

The full implementation is publicly available at \href{https://github.com/biubiutomato/TME-Agent}{https://github.com/biubiutomato/TME-Agent}.

\bibliographystyle{plain}
\clearpage
\section*{Appendix: Prompt Construction Details}

To illustrate how prompt token usage differs between methods, we include the full prompt content passed to the model at each interaction round. The Baseline approach appends the full history at each step, causing linear growth in prompt size. In contrast, the Task Execution Tree (TET) method reuses structured memory and generates minimal prompts based on task nodes. This architectural difference is the primary source of token efficiency observed in Table~\ref{tab:token_usage_prompt_completion_total}.

\begin{table}[ht]
\centering
\scriptsize
\renewcommand{\arraystretch}{1.3}
\begin{tabular}{|c|p{6.2cm}|p{6.2cm}|}
\hline
\textbf{Round} & \textbf{Baseline Prompt (Full History)} & \textbf{TET Prompt (Structured Memory)} \\
\hline
\textbf{Round 1} &
User: Help me fill out a form, I will provide some of my information to you.
&
User: Help me fill out a form, I will provide some of my information to you.
\\
\hline
\textbf{Round 2} &
User: Help me fill out a form, I will provide some of my information to you. \newline
Assistant: Of course! Please provide me with the information you would like me to include in the form. \newline
User: My name is John Doe.
&
User: Help me fill out a form, I will provide some of my information to you. \newline
Assistant: Of course! Please provide me with the information you would like me to include in the form. \newline
User: My name is John Doe.
\\
\hline
\textbf{Round 3} &
User: Help me fill out a form, I will provide some of my information to you. \newline
Assistant: Of course! Please provide me with the information you would like me to include in the form. \newline
User: My name is John Doe. \newline
Assistant: Great, thank you! What other information would you like me to include in the form. \newline
User: My email is john@example.com.
&
User: Help me fill out a form, I will provide some of my information to you. \newline
Assistant: Of course! Please provide me with the information you would like me to include in the form. \newline
User: My email is john@example.com.
\\
\hline
\textbf{Round 4} &
User: Help me fill out a form, I will provide some of my information to you. \newline
Assistant: Of course! Please provide me with the information you would like me to include in the form. \newline
User: My name is John Doe. \newline
Assistant: Great, thank you! What other information would you like me to include in the form. \newline
User: My email is john@example.com. \newline
Assistant: Great, thank you for sharing your email address. Can you please provide me with the rest of the information needed to fill out the form? \newline
User: My address is Market Street, San Francisco.
&
User: Help me fill out a form, I will provide some of my information to you. \newline
Assistant: Of course! Please provide me with the information you would like me to include in the form. \newline
User: My address is Market Street, San Francisco.
\\
\hline
\textbf{Round 5} &
User: Help me fill out a form, I will provide some of my information to you. \newline
Assistant: Of course! Please provide me with the information you would like me to include in the form. \newline
User: My name is John Doe. \newline
Assistant: Great, thank you! What other information would you like me to include in the form. \newline
User: My email is john@example.com. \newline
Assistant: Great, thank you for sharing your email address. Can you please provide me with the rest of the information needed to fill out the form? \newline
User: My address is Market Street, San Francisco. \newline
Assistant: Great, thank you for providing your address. Can you please provide me with your full name, date of birth, email address, and phone number so I can complete the form accurately? \newline
User: Sorry, to correct, my name is John Smith.
&
User: Help me fill out a form, I will provide some of my information to you. \newline
Assistant: Of course! Please provide me with the information you would like me to include in the form. \newline
User: My name is John Doe. \newline
Assistant: Great, thank you! What other information would you like me to include in the form. \newline
User: Sorry, to correct, my name is John Smith.
\\
\hline
\textbf{Round 6} &
User: Help me fill out a form, I will provide some of my information to you. \newline
Assistant: Of course! Please provide me with the information you would like me to include in the form. \newline
User: My name is John Doe. \newline
Assistant: Great, thank you! What other information would you like me to include in the form. \newline
User: My email is john@example.com. \newline
Assistant: Great, thank you for sharing your email address. Can you please provide me with the rest of the information needed to fill out the form? \newline
User: My address is Market Street, San Francisco. \newline
Assistant: Great, thank you for providing your address. Can you please provide me with your full name, date of birth, email address, and phone number so I can complete the form accurately? \newline
User: Sorry, to correct, my name is John Smith. \newline
Assistant: Got it, thank you for the correction. Please provide me with the rest of the information you would like me to include in the form. \newline
User: Help to repeat my information, Then submit.
&
User: Help me fill out a form, I will provide some of my information to you. \newline
Assistant: Of course! Please provide me with the information you would like me to include in the form. \newline
User: My name is John Doe.
Assistant: Great, thank you! What other information would you like me to include in the form. \newline
User: Sorry, to correct, my name is John Smith.
Assistant: Got it, thank you for the correction. Please provide me with the rest of the information you would like me to include in the form. \newline
User: My email is john@example.com. \newline
Assistant: Great, thank you for sharing your email address. Can you please provide me with the rest of the information needed to fill out the form? \newline
User: My address is Market Street, San Francisco. Assistant: Great, thank you for providing your ad- dress. Can you please provide me with your full name, date of birth, email address, and phone num- ber so I can complete the form accurately?
User: Help to repeat my information, Then submit.
\\
\hline
\end{tabular}
\caption{Prompt construction in each round for the Baseline and TET methods. TET’s structured memory design avoids redundant history, resulting in significant prompt token savings.}
\label{tab:prompt_construction_comparison}
\end{table}

\bibliography{refs}

\end{document}